\documentclass[11pt]{article}
\usepackage{acl2016}
\usepackage[font=small]{caption}
\usepackage{times}
\usepackage{latexsym}
\usepackage{amsmath}
\usepackage{multirow}
\usepackage{url}
\usepackage{graphicx}
\usepackage{anyfontsize}
\usepackage{color}
\aclfinalcopy 
\def\aclpaperid{289}
\DeclareMathOperator*{\argmax}{arg\,max}
\title{Learning-Based Single-Document Summarization with\\Compression and Anaphoricity Constraints}

\author{Greg Durrett \\
  Computer Science Division \\
  UC Berkeley \\
  {\fontsize{10}{12} {\tt gdurrett@cs.berkeley.edu}} \\\And
  Taylor Berg-Kirkpatrick \\
  School of Computer Science \\
  Carnegie Mellon University \\
  {\fontsize{10}{12} {\tt tberg@cs.cmu.edu}} \\\And
  Dan Klein \\
  Computer Science Division \\
  UC Berkeley\\
  {\fontsize{10}{12} {\tt klein@cs.berkeley.edu}}}

\date{}

\begin{document}
\maketitle
\begin{abstract}
We present a discriminative model for single-document summarization that integrally combines compression and anaphoricity constraints.
Our model selects textual units to include in the summary based on a rich set of sparse features whose weights are learned on a large corpus.
We allow for the deletion of content within a sentence when that deletion is licensed by compression rules; in our framework, these are implemented as dependencies between subsentential units of text.
Anaphoricity constraints then improve cross-sentence coherence by guaranteeing that, for each pronoun included in the summary, the pronoun's antecedent is included as well or the pronoun is rewritten as a full mention.
When trained end-to-end, our final system\footnote{Available at \texttt{http://nlp.cs.berkeley.edu}} outperforms prior work on both ROUGE as well as on human judgments of linguistic quality.
\end{abstract}

\section{Introduction}

While multi-document summarization is well-studied in the NLP literature \cite{CarbonellGoldstein1998,GillickFavre2009,LinBilmes2011,NenkovaMcKeown2011}, single-document summarization \cite{McKeownEtAl1995,Marcu1998,Mani2001,HiraoEtAl2013} has received less attention in recent years and is generally viewed as more difficult.
Content selection is tricky without redundancy across multiple input documents as a guide and simple positional information is often hard to beat \cite{PennZhu2008}. In this work, we tackle the single-document problem by training an expressive summarization model on a large naturally occurring corpus---the New York Times Annotated Corpus \cite{Sandhaus2008} which contains around 100,000 news articles with abstractive summaries---learning to select important content with lexical features. This corpus has been explored in related contexts \cite{DunietzGillick2014,HongNenkova2014}, but to our knowledge it has not been directly used for single-document summarization.


\begin{figure*}[t!]
\begin{centering}
\includegraphics[scale=0.70]{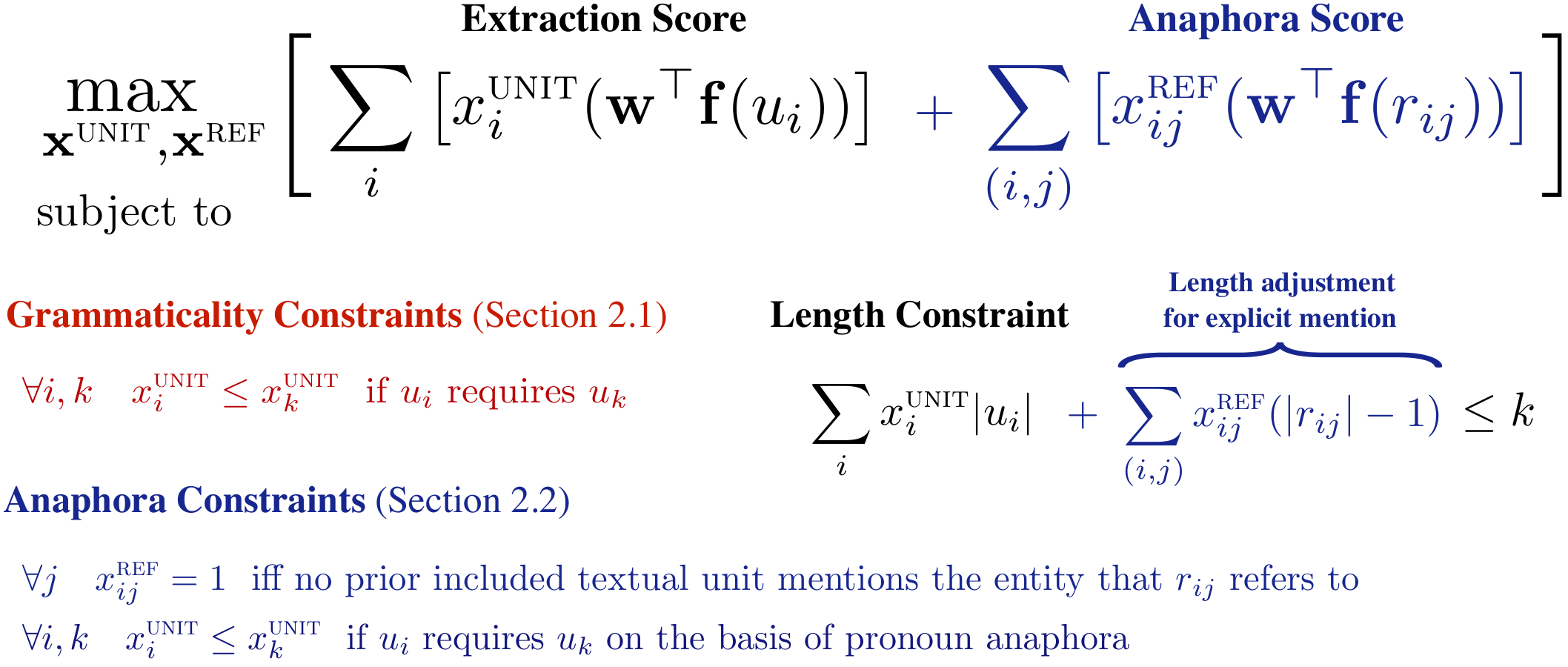}
\caption{\label{fig:model} ILP formulation of our single-document summarization model.
The basic model extracts a set of textual units with binary variables $\mathbf{x}^\textrm{\textsc{unit}}$ subject to a length constraint.
These textual units $\mathbf{u}$ are scored with weights $\mathbf{w}$ and features $\mathbf{f}$.
Next, we add constraints derived from both syntactic parses and Rhetorical Structure Theory (RST) to enforce grammaticality.
Finally, we add anaphora constraints derived from coreference in order to improve summary coherence.
We introduce additional binary variables $\mathbf{x}^\textrm{\textsc{ref}}$ that control whether each pronoun is replaced with its antecedent using a candidate replacement $r_{ij}$.
These are also scored in the objective and are incorporated into the length constraint.
}
\end{centering}
\end{figure*}



To increase the expressive capacity of our model we allow more aggressive compression of individual sentences by combining two different formalisms---one syntactic and the other discursive. Additionally, we incorporate a model of anaphora resolution and give our system the ability rewrite pronominal mentions, further increasing expressivity.
In order to guide the model, we incorporate (1) constraints from coreference ensuring that critical pronoun references are clear in the final summary and (2) constraints from syntactic and discourse parsers ensuring that sentence realizations are well-formed.
Despite the complexity of these additional constraints, we demonstrate an efficient inference procedure using an ILP-based approach.
By training our full system end-to-end on a large-scale dataset, we are able to learn a high-capacity structured model of the summarization process, contrasting with past approaches to the single-document task which have typically been heuristic in nature \cite{DaumeMarcu2002,HiraoEtAl2013}.

We focus our evaluation on the New York Times Annotated corpus \cite{Sandhaus2008}.
According to ROUGE, our system outperforms a document prefix baseline, a bigram coverage baseline adapted from a strong multi-document system \cite{GillickFavre2009}, and a discourse-informed method from prior work \cite{YoshidaEtAl2014}.
Imposing discursive and referential constraints improves human judgments of linguistic clarity and referential structure---outperforming the method of \newcite{YoshidaEtAl2014} and approaching the clarity of a sentence-extractive baseline---and still achieves substantially higher ROUGE score than either method.
These results indicate that our model has the expressive capacity to extract important content, but is sufficiently constrained to ensure fluency is not sacrificed as a result.

Past work has explored various kinds of structure for summarization.
Some work has focused on improving content selection using discourse structure \cite{LouisEtAl2010,HiraoEtAl2013}, topical structure \cite{BarzilayLee2004}, or related techniques \cite{MithunKosseim2011}.
Other work has used structure primarily to reorder summaries and ensure coherence \cite{BarzilayEtAl2001,BarzilayLapata2008,LouisNenkova2012,ChristensenEtAl2013} or to represent content for sentence fusion or abstraction \cite{ThadaniMcKeown2013,PighinEtAl2014}.
Similar to these approaches, we appeal to structures from upstream NLP tasks (syntactic parsing, RST parsing, and coreference) to restrict our model's capacity to generate.
However, we go further by optimizing for ROUGE subject to these constraints with end-to-end learning.

\begin{figure*}[t!]
\begin{centering}
\includegraphics[trim=0mm 127mm 00mm 28mm,scale=0.89]{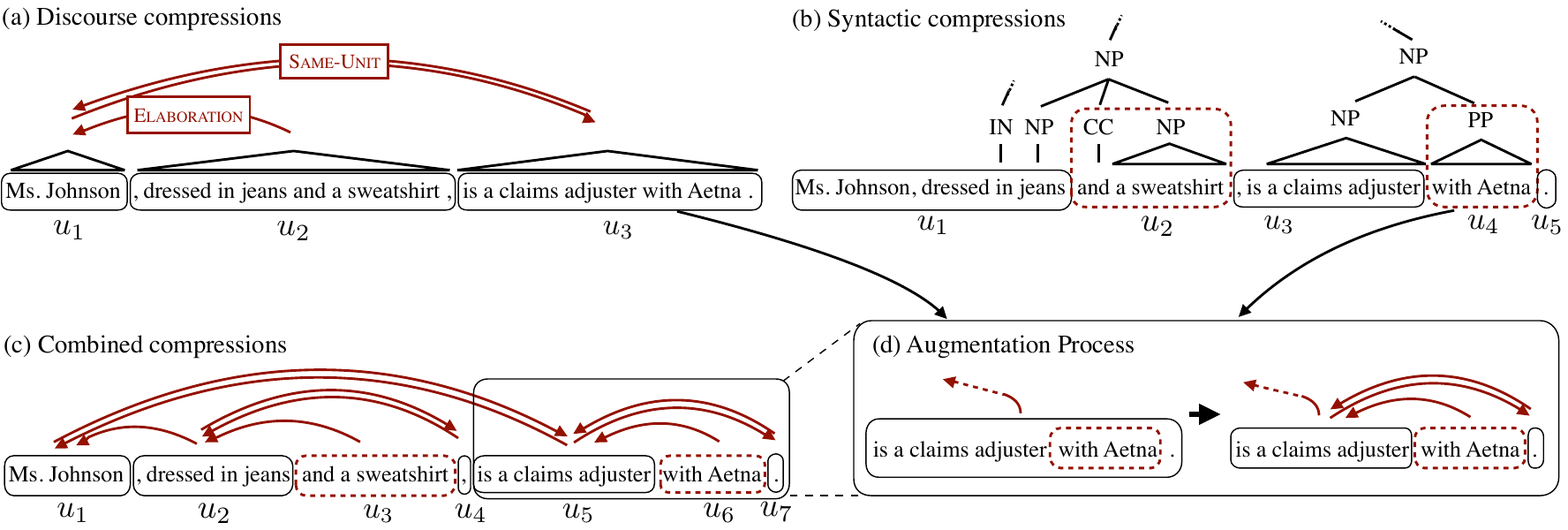}
\caption{\label{fig:compression} Compression constraints on an example sentence. (a) RST-based compression structure like that in \protect\newcite{HiraoEtAl2013}, where we can delete the \textsc{Elaboration} clause. (b) Two syntactic compression options from \protect\newcite{BergKirkpatrickEtAl2011}, namely deletion of a coordinate and deletion of a PP modifier. (c) Textual units and requirement relations (arrows) after merging all of the available compressions. (d) Process of augmenting a textual unit with syntactic compressions.
}
\end{centering}
\end{figure*}

\section{Model}
\label{sec:model}

Our model is shown in Figure~\ref{fig:model}.
Broadly, our ILP takes a set of textual units $\mathbf{u} = (u_1,\ldots,u_n)$ from a document and finds the highest-scoring extractive summary by optimizing over variables $\mathbf{x}^{\textrm{\textsc{unit}}} = x^{\textrm{\textsc{unit}}}_1,\ldots,x^{\textrm{\textsc{unit}}}_n$, which are binary indicators of whether each unit is included.
Textual units are contiguous parts of sentences that serve as the fundamental units of extraction in our model.
For a sentence-extractive model, these would be entire sentences, but for our compressive models we will have more fine-grained units, as shown in Figure~\ref{fig:compression} and described in Section~\ref{sec:grammaticality}.
Textual units are scored according to features $\mathbf{f}$ and model parameters $\mathbf{w}$ learned on training data.
Finally, the extraction process is subject to a length constraint of $k$ words.
This approach is similar in spirit to ILP formulations of multi-document summarization systems, though in those systems content is typically modeled in terms of bigrams \cite{GillickFavre2009,BergKirkpatrickEtAl2011,HongNenkova2014,LiEtAl2015}.
For our model, type-level $n$-gram scoring only arises when we compute our loss function in max-margin training (see Section~\ref{sec:learning}).

In Section~\ref{sec:grammaticality}, we discuss grammaticality constraints, which take the form of introducing dependencies between textual units, as shown in Figure~\ref{fig:compression}.
If one textual unit \emph{requires} another, it cannot be included unless its prerequisite is.
We will show that different sets of requirements can capture both syntactic and discourse-based compression schemes.

Furthermore, we introduce anaphora constraints (Section~\ref{sec:anaphora}) via a new set of variables that capture the process of rewriting pronouns to make them explicit mentions. That is, $x^{\textrm{\textsc{ref}}}_{ij}$ = 1 if we should rewrite the $j$th pronoun in the $i$th unit with its antecedent.
These pronoun rewrites are scored in the objective and introduced into the length constraint to make sure they do not cause our summary to be too long.
Finally, constraints on these variables control when they are used and also require the model to include antecedents of pronouns when the model is not confident enough to rewrite them.

\subsection{Grammaticality Constraints}
\label{sec:grammaticality}

Following work on isolated sentence compression \cite{McDonald2006,ClarkeLapata2008} and compressive summarization \cite{Lin2003,MartinsSmith2009,BergKirkpatrickEtAl2011,WoodsendLapata2012,AlmeidaMartins2013}, we wish to be able to compress sentences so we can pack more information into a summary.
During training, our model learns how to take advantage of available compression options and select content to match human generated summaries as closely possible.\footnote{The features in our model are actually rich enough to learn a sophisticated compression model, but the data we have (abstractive summaries) does not directly provide examples of correct compressions; past work has gotten around this with multi-task learning \cite{AlmeidaMartins2013}, but we simply treat grammaticality as a constraint from upstream models.} We explore two ways of deriving units for compression: the RST-based compressions of \newcite{HiraoEtAl2013} and the syntactic compressions of \newcite{BergKirkpatrickEtAl2011}.

\paragraph{RST compressions} Figure~\ref{fig:compression}a shows how to derive compressions from Rhetorical Structure Theory \cite{MannThompson1988,CarlsonEtAl2001}.
We show a sentence broken into elementary discourse units (EDUs) with RST relations between them.
Units marked as \textsc{Same-Unit} must both be kept or both be deleted, but other nodes in the tree structure can be deleted as long as we do not delete the parent of an included node.
For example, we can delete the \textsc{Elaboration} clause, but we can delete neither the first nor last EDU.
Arrows depict the constraints this gives rise to in the ILP (see Figure~\ref{fig:model}): $u_2$ requires $u_1$, and $u_1$ and $u_3$ mutually require each other.
This is a more constrained form of compression than was used in past work \cite{HiraoEtAl2013}, but we find that it improves human judgments of fluency (Section~\ref{sec:nyt_results}).

\paragraph{Syntactic compressions} Figure~\ref{fig:compression}b shows two examples of compressions arising from syntactic patterns \cite{BergKirkpatrickEtAl2011}: deletion of the second part of a coordinated NP and deletion of a PP modifier to an NP.
These patterns were curated to leave sentences as grammatical after being compressed, though perhaps with damaged semantic content.

\paragraph{Combined compressions} Figure~\ref{fig:compression}c shows the textual units and requirement relations yielded by combining these two types of compression.
On this example, the two schemes capture orthogonal compressions, and more generally we find that they stack to give better results for our final system (see Section~\ref{sec:nyt_results}).
To actually synthesize textual units and the constraints between them, we start from the set of RST textual units and introduce syntactic compressions as new children when they don't cross existing brackets; because syntactic compressions are typically narrower in scope, they are usually completely contained in EDUs.
Figure~\ref{fig:compression}d shows an example of this process: the possible deletion of \emph{with Aetna} is grafted onto the textual unit and appropriate requirement relations are introduced.
The net effect is that the textual unit is wholly included, partially included (\emph{with Aetna} removed), or not at all.

Formally, we define an RST tree as $T_\textrm{rst} = (S_\textrm{rst}, \pi_\textrm{rst})$ where $S_\textrm{rst}$ is a set of EDU spans $(i,j)$ and $\pi : S \rightarrow 2^S$ is a mapping from each EDU span to EDU spans it depends on.
Syntactic compressions can be expressed in a similar way with trees $T_\textrm{syn}$.
These compressions are typically smaller-scale than EDU-based compressions, so we use the following modification scheme.
Denote by $T_{\textrm{syn}(kl)}$ a nontrivial (supports some compression) subtree of $T_\textrm{syn}$ that is completely contained in an EDU $(i,j)$. We build the following combined compression tree, which we refer to as the \emph{augmentation} of $T_\textrm{rst}$ with $T_{\textrm{syn}(kl)}$:

\vspace{-0.15in}
\small
\begin{align*}
T_\textrm{comb} &= (S \cup S_\textrm{syn}(kl) \cup \{(i,k),(l,j)\}, \pi_\textrm{rst} \cup \pi_{\textrm{syn}(kl)} \cup \\
   &\{(i,k) \rightarrow (l,j), (l,j) \rightarrow (i,k), (k,l) \rightarrow (i,k)\})
\end{align*}
\normalsize
That is, we maintain the existing tree structure except for the EDU $(i,j)$, which is broken into three parts: the outer two depend on each other (\emph{is a claims adjuster} and \emph{.} from Figure~\ref{fig:compression}d) and the inner one depends on the others and preserves the tree structure from $T_\textrm{syn}$.
We augment $T_\textrm{rst}$ with all maximal subtrees of $T_\textrm{syn}$, i.e.~all trees that are not contained in other trees that are used in the augmentation process.

This is broadly similar to the combined compression scheme in \newcite{KikuchiEtAl2014} but we use a different set of constraints that more strictly enforce grammaticality.\footnote{We also differ from past work in that we do not use cross-sentential RST constraints \cite{HiraoEtAl2013,YoshidaEtAl2014}. We experimented with these and found no improvement from using them, possibly because we have a feature-based model rather than a heuristic content selection procedure, and possibly because automatic discourse parsers are less good at recovering cross-sentence relations.}

\begin{figure}[t!]
\begin{centering}
\includegraphics[trim=-5mm 75mm 0mm 0mm,scale=0.57]{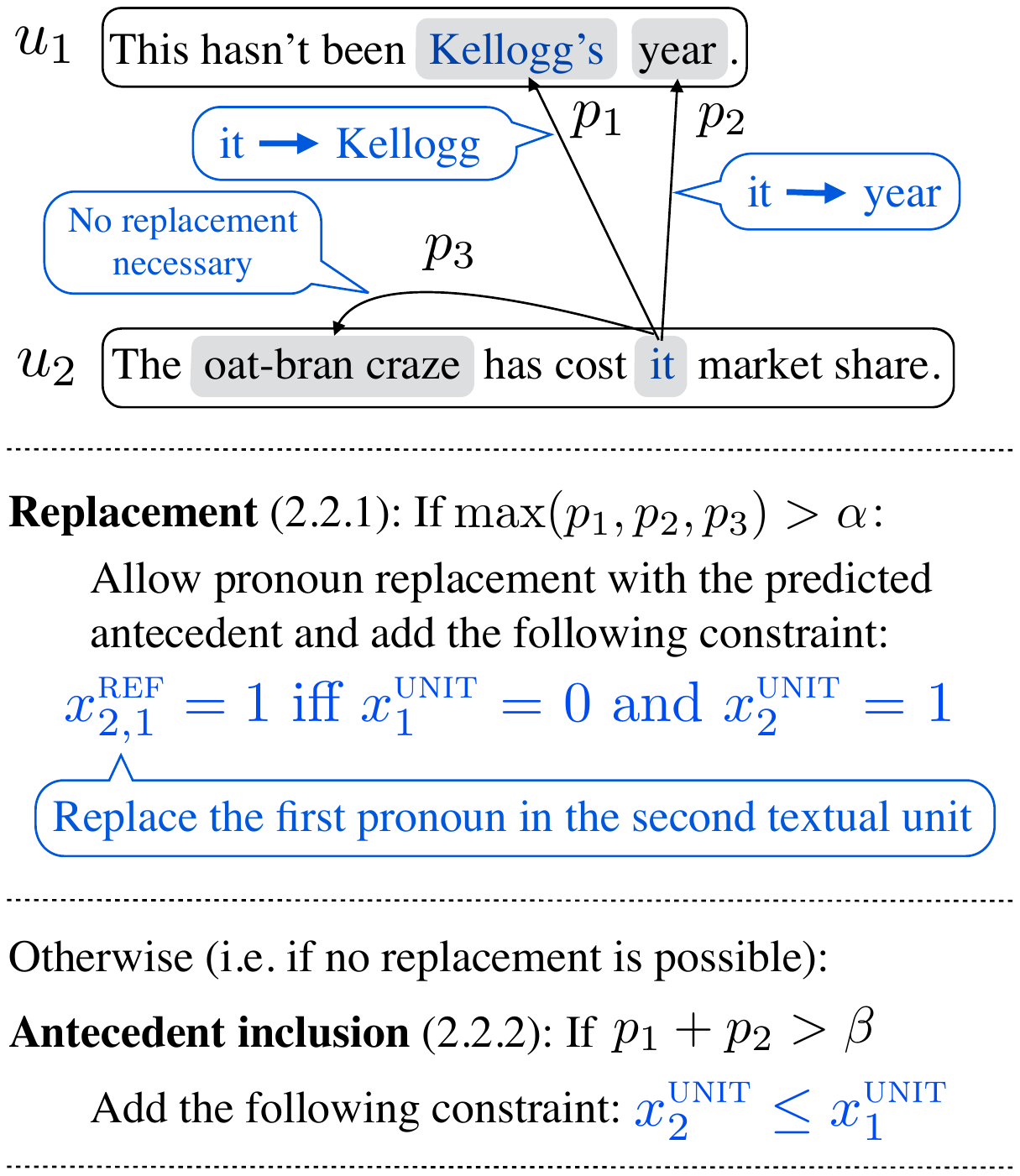}
\caption{\label{fig:pronouns} Modifications to the ILP to capture pronoun coherence. \emph{It}, which refers to \emph{Kellogg}, has several possible antecedents from the standpoint of an automatic coreference system \cite{DurrettKlein2014}. If the coreference system is confident about its selection (above a threshold $\alpha$ on the posterior probability), we allow for the model to explicitly replace the pronoun if its antecedent would be deleted (Section~\ref{sec:pron_rep}). Otherwise, we merely constrain one or more probable antecedents to be included (Section~\ref{sec:pron_ant}); even if the coreference system is incorrect, a human can often correctly interpret the pronoun with this additional context.
}
\end{centering}
\end{figure}

\subsection{Anaphora Constraints}
\label{sec:anaphora}

What kind of cross-sentential coherence do we need to ensure for the kinds of summaries our system produces?
Many notions of coherence are useful, including centering theory \cite{GroszEtAl1995} and lexical cohesion \cite{NishikawaEtAl2014}, but one of the most pressing phenomena to deal with is pronoun anaphora \cite{ClarkeLapata2010}.
Cases of pronouns being ``orphaned'' during extraction (their antecedents are deleted) are relatively common: they occur in roughly 60\% of examples produced by our summarizer when no anaphora constraints are enforced.
This kind of error is particularly concerning for summary interpretation and impedes the ability of summaries to convey information effectively \cite{Grice1975}.
Our solution is to explicitly impose constraints on the model based on pronoun anaphora resolution.\footnote{We focus on pronoun coreference because it is the most pressing manifestation of this problem and because existing coreference systems perform well on pronouns compared to harder instances of coreference \cite{DurrettKlein2013}.}

Figure~\ref{fig:pronouns} shows an example of a problem case.
If we extract only the second textual unit shown, the pronoun \emph{it} will lose its antecedent, which in this case is \emph{Kellogg}.
We explore two types of constraints for dealing with this: rewriting the pronoun explicitly, or constraining the summary to include the pronoun's antecedent.

\subsubsection{Pronoun Replacement}
\label{sec:pron_rep}

One way of dealing with these pronoun reference issues is to explicitly replace the pronoun with what it refers to.
This replacement allows us to maintain maximal extraction flexibility, since we can make an isolated textual unit meaningful even if it contains a pronoun.
Figure~\ref{fig:pronouns} shows how this process works.
We run the Berkeley Entity Resolution System \cite{DurrettKlein2014} and compute posteriors over possible links for the pronoun.
If the coreference system is sufficiently confident in its prediction (i.e.~$\max_i p_i > \alpha$ for a specified threshold $\alpha > \frac{1}{2}$), we allow ourselves to replace the pronoun with the first mention of the entity corresponding to the pronoun's most likely antecedent.
In Figure~\ref{fig:pronouns}, if the system correctly determines that \emph{Kellogg} is the correct antecedent with high probability, we enable the first replacement shown there, which is used if $u_2$ is included the summary without $u_1$.\footnote{If the proposed replacement is a proper mention, we replace the pronoun just with the subset of the mention that constitutes a named entity (rather than the whole noun phrase). We control for possessive pronouns by deleting or adding \emph{'s} as appropriate.}

As shown in the ILP in Figure~\ref{fig:model}, we instantiate corresponding pronoun replacement variables $\mathbf{x}^\textrm{\textsc{ref}}$ where $x^\textrm{\textsc{ref}}_{ij} = 1$ implies that the $j$th pronoun in the $i$th sentence should be replaced in the summary.
We use a candidate pronoun replacement if and only if the pronoun's corresponding (predicted) entity hasn't been mentioned previously in the summary.\footnote{Such a previous mention may be a pronoun; however, note that that pronoun would then be targeted for replacement unless its antecedent were included somehow.}
Because we are generally replacing pronouns with longer mentions, we also need to modify the length constraint to take this into account.
Finally, we incorporate features on pronoun replacements in the objective, which helps the model learn to prefer pronoun replacements that help it to more closely match the human summaries.

\subsubsection{Pronoun Antecedent Constraints}
\label{sec:pron_ant}

Explicitly replacing pronouns is risky: if the coreference system makes an incorrect prediction, the intended meaning of the summary may be damaged.
Fortunately, the coreference model's posterior probabilities have been shown to be well-calibrated \cite{NguyenOConnor2015}, meaning that cases where it is likely to make errors are signaled by flatter posterior distributions.
In this case, we enable a more conservative set of constraints that include additional content in the summary to make the pronoun reference clear without explicitly replacing it.
This is done by requiring the inclusion of any textual unit which contains possible pronoun references whose posteriors sum to at least a threshold parameter $\beta$.
Figure~\ref{fig:pronouns} shows that this constraint can force the inclusion of $u_1$ to provide additional context.
Although this could still lead to unclear pronouns if text is stitched together in an ambiguous or even misleading way, in practice we observe that the textual units we force to be added almost always occur very recently before the pronoun, giving enough additional context for a human reader to figure out the pronoun's antecedent unambiguously.

\subsection{Features}
\label{sec:features}

The features in our model (see Figure~\ref{fig:model}) consist of a set of surface indicators capturing mostly lexical and configurational information. Their primary role is to identify important document content. The first three types of features fire over textual units, the last over pronoun replacements.

\paragraph{Lexical} These include indicator features on non-stopwords in the textual unit that appear at least five times in the training set and analogous POS features. We also use lexical features on the first, last, preceding, and following words for each textual unit. Finally, we conjoin each of these features with an indicator of bucketed position in the document (the index of the sentence containing the textual unit).

\paragraph{Structural} These features include various conjunctions of the position of the textual unit in the document, its length, the length of its corresponding sentence, the index of the paragraph it occurs in, and whether it starts a new paragraph (all values are bucketed).

\paragraph{Centrality} These features capture rough information about the centrality of content: they consist of bucketed word counts conjoined with bucketed sentence index in the document. We also fire features on the number of times of each entity mentioned in the sentence is mentioned in the rest of the document (according to a coreference system), the number of entities mentioned in the sentence, and surface properties of mentions including type and length

\paragraph{Pronoun replacement} These target properties of the pronoun replacement such as its length, its sentence distance from the current mention, its type (nominal or proper), and the identity of the pronoun being replaced.

\section{Learning}
\label{sec:learning}

We learn weights $\mathbf{w}$ for our model by training on a large corpus of documents $\mathbf{u}$ paired with reference summaries $\mathbf{y}$. We formulate our learning problem as a standard instance of structured SVM (see \newcite{Smith2011} for an introduction). Because we want to optimize explicitly for ROUGE-1,\footnote{We found that optimizing for ROUGE-1 actually resulted in a model with better performance on both ROUGE-1 and ROUGE-2. We hypothesize that this is because framing our optimization in terms of ROUGE-2 would lead to a less nuanced set of constraints: bigram matches are relatively rare when the reference is a short, abstractive summary, so a loss function based on ROUGE-2 will express a flatter preference structure among possible outputs.} we define a ROUGE-based loss function that accommodates the nature of our supervision, which is in terms of abstractive summaries $\mathbf{y}$ that in general cannot be produced by our model. Specifically, we take:

\vspace{-0.1in}
\small
\begin{equation*}
\resizebox{1.0\hsize}{!}{$\ell(\mathbf{x}^\textrm{\textsc{ngram}},\mathbf{y}) = \max_{\mathbf{x^*}} \textrm{\textsc{rouge-1}}(\mathbf{x^*},\mathbf{y}) - \textrm{\textsc{rouge-1}}(\mathbf{x}^\textrm{\textsc{ngram}},\mathbf{y})$}
\end{equation*}
\normalsize
i.e.~the gap between the hypothesis's ROUGE score and the oracle ROUGE score achievable under the model (including constraints). Here $\mathbf{x}^\textrm{\textsc{ngram}}$ are indicator variables that track, for each $n$-gram type in the reference summary, whether that $n$-gram is present in the system summary. These are the sufficient statistics for computing ROUGE.

We train the model via stochastic subgradient descent on the primal form of the structured SVM objective \cite{RatliffEtAl2007,KummerfeldEtAl2015}. In order to compute the subgradient for a given training example, we need to find the most violated constraint on the given instance through a loss-augmented decode, which for a linear model takes the form $\argmax_\mathbf{x} \mathbf{w}^\top \mathbf{f}(\mathbf{x}) + \ell(\mathbf{x},\mathbf{y})$. To do this decode at training time in the context of our model, we use an extended version of our ILP in Figure~\ref{fig:model} that is augmented to explicitly track type-level $n$-grams:


%


\vspace{-0.2in}
\small
\begin{align*}
&\max_{\mathbf{x}^\textrm{\textsc{unit}},\mathbf{x}^\textrm{\textsc{ref}},\mathbf{x}^\textrm{\textsc{ngram}}} \left[ \sum_i \left[ x_i^\textrm{\textsc{unit}} (\mathbf{w}^\top \mathbf{f}(u_i))\right] \right. \\
&\ \ \ \ \ \ \ \ \ \ \ \ \ \left. + \sum_{(i,j)} \left[ x_{ij}^\textrm{\textsc{ref}} (\mathbf{w}^\top \mathbf{f}(r_{ij}))\right] - \ell(\mathbf{x}^\textrm{\textsc{ngram}},\mathbf{y}) \right] \\
&\textrm{subject to all constraints from Figure~\ref{fig:model}, and}\\
&x_i^\textrm{\textsc{ngram}} = 1\ \textrm{iff an included textual unit or replacement}\\
&\ \ \ \ \ \ \ \ \ \ \ \ \ \ \ \ \ \ \textrm{ contains the $i$th reference $n$-gram}
\end{align*}
\normalsize
These kinds of variables and constraints are common in multi-document summarization systems that score bigrams (Gillick and Favre, 2009 \emph{inter alia})\nocite{GillickFavre2009}. Note that since ROUGE is only computed over non-stopword $n$-grams and pronoun replacements only replace pronouns, pronoun replacement can never remove an $n$-gram that would otherwise be included. 



For all experiments, we optimize our objective using AdaGrad \cite{DuchiEtAl2011} with $\ell_1$ regularization ($\lambda = 10^{-8}$, chosen by grid search), with a step size of 0.1 and a minibatch size of 1. We train for 10 iterations on the training data, at which point held-out model performance no longer improves. Finally, we set the anaphora thresholds $\alpha = 0.8$ and $\beta = 0.6$ (see Section~\ref{sec:anaphora}). The values of these and other hyperparameters were determined on a held-out development set from our New York Times training data. All ILPs are solved using GLPK version 4.55.


\section{Experiments}
\label{sec:experiments}

We primarily evaluate our model on a roughly 3000-document evaluation set from the New York Times Annotated Corpus \cite{Sandhaus2008}. We also investigate its performance on the RST Discourse Treebank \cite{CarlsonEtAl2001}, but because this dataset is only 30 documents it provides much less robust estimates of performance.\footnote{Tasks like DUC and TAC have focused on multi-document summarization since around 2003, hence the lack of more standard datasets for single-document summarization.} Throughout this section, when we decode a document, we set the word budget for our summarizer to be the same as the number of words in the corresponding reference summary, following previous work \cite{HiraoEtAl2013,YoshidaEtAl2014}.

\begin{figure*}[t!]
\begin{centering}
\includegraphics[scale=0.60]{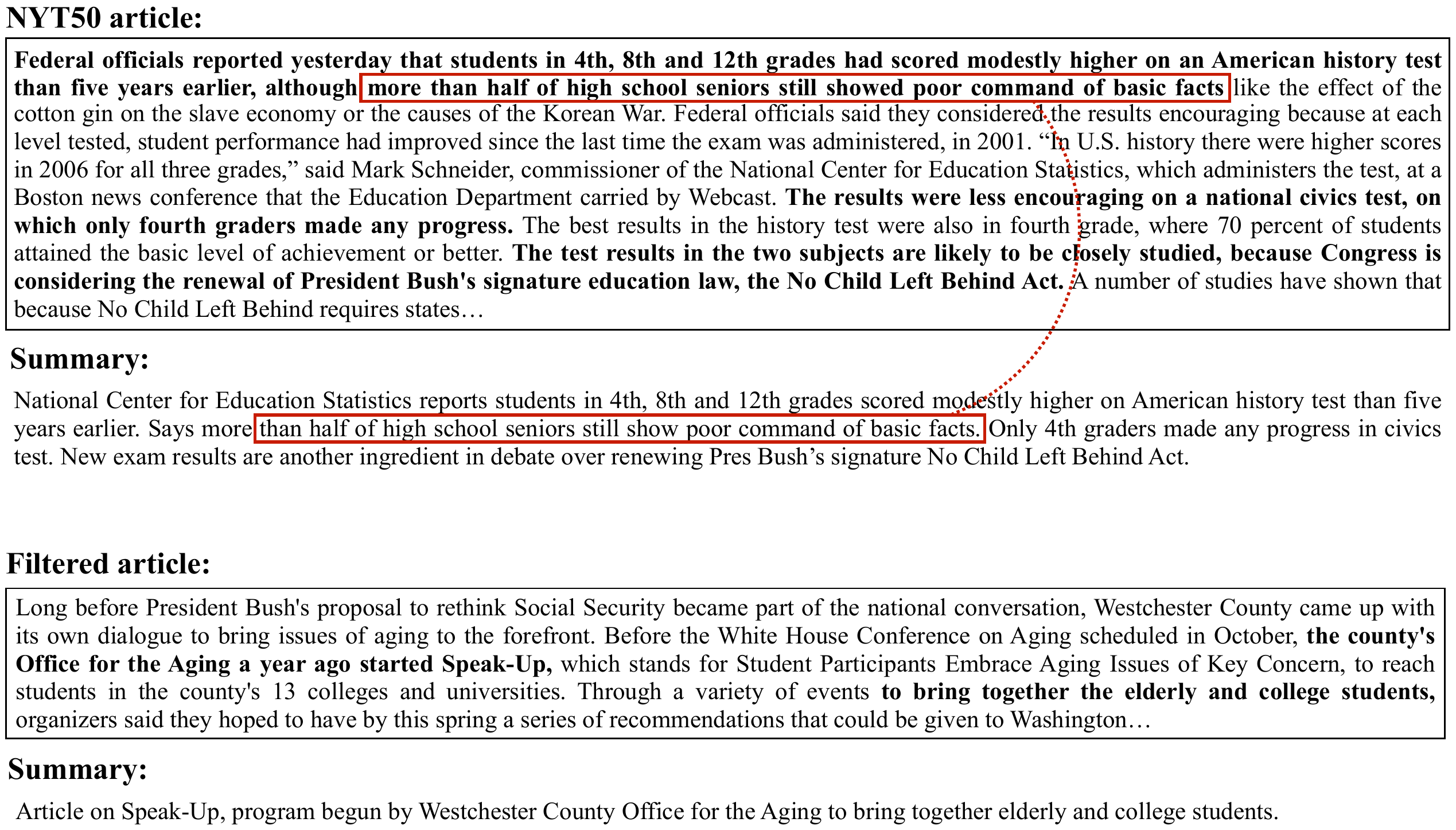}
\caption{\label{fig:example} Examples of an article kept in the NYT50 dataset (top) and an article removed because the summary is too short. The top summary has a rich structure to it, corresponding to various parts of the document (bolded) and including some text that is essentially a direct extraction.
}
\end{centering}
\end{figure*}

\subsection{Preprocessing}
\label{sec:preprocessing}

We preprocess all data using the Berkeley Parser \cite{PetrovEtAl2006}, specifically the GPU-accelerated version of the parser from \newcite{HallEtAl2014}, and the Berkeley Entity Resolution System \cite{DurrettKlein2014}. For RST discourse analysis, we segment text into EDUs using a semi-Markov CRF trained on the RST treebank with features on boundaries similar to those of \newcite{HernaultEtAl2010}, plus novel features on spans including span length and span identity for short spans.

To follow the conditions of \newcite{YoshidaEtAl2014} as closely as possible, we also build a discourse parser in the style of \newcite{HiraoEtAl2013}, since their parser is not publicly available. Specifically, we use the first-order projective parsing model of \newcite{McDonaldEtAl2005} and features from \newcite{SoricutMarcu2003}, \newcite{HernaultEtAl2010}, and \newcite{JotyEtAl2013}.
When using the same head annotation scheme as \newcite{YoshidaEtAl2014}, we outperform their discourse dependency parser on unlabeled dependency accuracy, getting 56\% as opposed to 53\%.

\subsection{New York Times Corpus}
\label{sec:nyt_corpus}

\begin{figure}[t!]
\begin{centering}
\includegraphics[trim=1mm 140mm 80mm 0mm,scale=0.61]{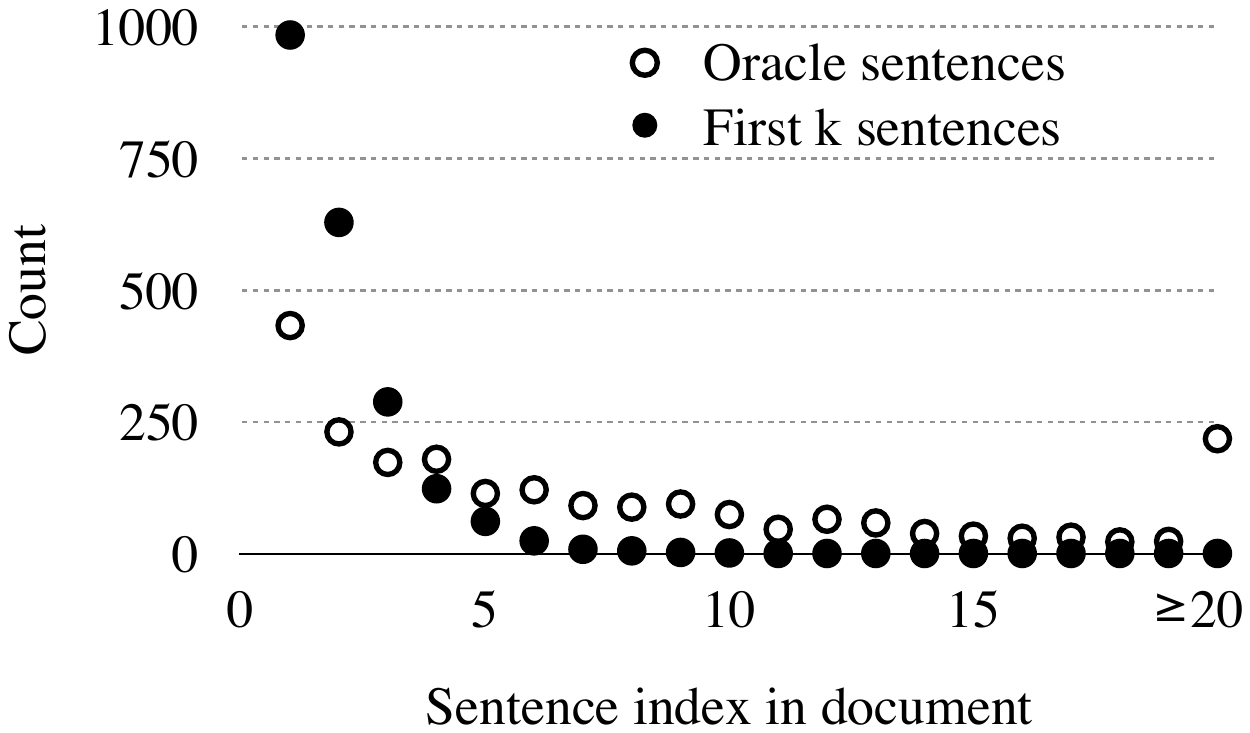}
\caption{\label{fig:nyt-graph} Counts on a 1000-document sample of how frequently both a document prefix baseline and a ROUGE oracle summary contain sentences at various indices in the document. There is a long tail of useful sentences later in the document, as seen by the fact that the oracle sentence counts drop off relatively slowly. Smart selection of content therefore has room to improve over taking a prefix of the document.
}
\vspace{-0.1in}
\end{centering}
\end{figure}

We now provide some details about the New York Times Annotated corpus. This dataset contains 110,540 articles with abstractive summaries; we split these into 100,834 training and 9706 test examples, based on date of publication (test is all articles published on January 1, 2007 or later). Examples of two documents from this dataset are shown in Figure~\ref{fig:example}. The bottom example demonstrates that some summaries are extremely short and formulaic (especially those for obituaries and editorials). To counter this, we filter the raw dataset by removing all documents with summaries that are shorter than 50 words. One benefit of filtering is that the length distribution of our resulting dataset is more in line with standard summarization evaluations like DUC; it also ensures a sufficient number of tokens in the budget to produce nontrivial summaries.
The filtered test set, which we call NYT50, includes 3,452 test examples out of the original 9,706.

Interestingly, this dataset is one where the classic document prefix baseline can be substantially outperformed, unlike in some other summarization settings \cite{PennZhu2008}. We show this fact explicitly in Section~\ref{sec:nyt_results}, but Figure~\ref{fig:nyt-graph} provides additional analysis in this regard. We compute oracle ROUGE-1 sentence-extractive summaries on a 1000-document subset of the training set and look at where the extracted sentences lie in the document. While they certainly skew earlier in the document, they do not all fall within the document prefix summary. One reason for this is that many of the articles are longer-form pieces that begin with a relatively content-free lede of several sentences, which should be identifiable with lexicosyntactic indicators as are used in our discriminative model.


\subsection{New York Times Results}
\label{sec:nyt_results}

\begin{table}[t]
\begin{center}
\footnotesize
\renewcommand{\tabcolsep}{1.5mm}
\begin{tabular}{|r|rr|rr|} \hline
 & R-1 $\uparrow$ & R-2 $\uparrow$ & CG $\uparrow$ & UP $\downarrow$ \\ \hline
\multicolumn{5}{|c|}{Baselines} \\ \hline
First sentences & 28.6 & 17.3   & 8.21 & 0.28 \\ 
First $k$ words & 35.7 & 21.6   & $-$ & $-$ \\ 
Bigram Frequency& 25.1 & 9.8   & $-$ & $-$ \\\hline
\multicolumn{5}{|c|}{Past work} \\ \hline
Tree Knapsack   & 34.7 & 19.6   & 7.20 & 0.42 \\ \hline
\multicolumn{5}{|c|}{This work} \\ \hline
Sentence extraction & 38.8 & 23.5     & 7.93 & 0.32 \\ 
EDU extraction      & 41.9 & 25.3     & 6.38 & 0.65 \\ 
Full   & \textbf{42.2} & \textbf{25.9}     & *$\dagger$7.52 & *0.36 \\ \hline
\multicolumn{5}{|c|}{Ablations from Full} \\ \hline
No Anaphoricity & 42.5 & 26.3     & 7.46 & 0.44 \\
No Syntactic Compr & 41.1 & 25.0    & $-$ & $-$ \\
No Discourse Compr & 40.5 & 24.7    & $-$ & $-$ \\ \hline
\end{tabular}
\end{center}
\caption{\label{table:nyt_test} Results on the NYT50 test set (documents with summaries of at least 50 tokens) from the New York Times Annotated Corpus \protect\cite{Sandhaus2008}. We report ROUGE-1 (R-1), ROUGE-2 (R-2), clarity/grammaticality (CG), and number of unclear pronouns (UP) (lower is better). On content selection, our system substantially outperforms all baselines, our implementation of the tree knapsack system \protect\cite{YoshidaEtAl2014}, and learned extractive systems with less compression, even an EDU-extractive system that sacrifices grammaticality. On clarity metrics, our final system performs nearly as well as sentence-extractive systems. The symbols * and $\dagger$ indicate statistically significant gains compared to No Anaphoricity and Tree Knapsack (respectively) with $p < 0.05$ according to a bootstrap resampling test. We also see that removing either syntactic or EDU-based compressions decreases ROUGE.
}
\end{table}

We evaluate our system along two axes: first, on content selection, using ROUGE\footnote{We use the ROUGE 1.5.5 script with the following command line arguments: \texttt{-n 2 -x -m -s}. All given results are macro-averaged recall values over the test set.} \cite{LinHovy2003}, and second, on clarity of language and referential structure, using annotators from Amazon Mechanical Turk. We follow the method of \newcite{GillickLiu2010} for this evaluation and ask Turkers to rate a summary on how grammatical it is using a 10-point Likert scale. Furthermore, we ask how many unclear pronouns references there were in the text. The Turkers do not see the original document or the reference summary, and rate each summary in isolation. \newcite{GillickLiu2010} showed that for linguistic quality judgments (as opposed to content judgments), Turkers reproduced the ranking of systems according to expert judgments.

To speed up preprocessing and training time on this corpus, we further restrict our training set to only contain documents with fewer than 100 EDUs.
All told, the final system takes roughly 20 hours to make 10 passes through the subsampled training data (22,000 documents) on a single core of an Amazon EC2 r3.4xlarge instance.

Table~\ref{table:nyt_test} shows the results on the NYT50 corpus. We compare several variants of our system and baselines. For baselines, we use two variants of first $k$: one which must stop on a sentence boundary (which gives better linguistic quality) and one which always consumes $k$ tokens (which gives better ROUGE). We also use a heuristic sentence-extractive baseline that maximizes the document counts (term frequency) of bigrams covered by the summary, similar in spirit to the multi-document method of \newcite{GillickFavre2009}.\footnote{Other heuristic multi-document approaches could be compared to, e.g.~\newcite{HeEtAl2012}, but a simple term frequency method suffices to illustrate how these approaches can underperform in the single-document setting.} We also compare to our implementation of the Tree Knapsack method of \newcite{YoshidaEtAl2014}, which matches their results very closely on the RST Discourse Treebank when discourse trees are controlled for. Finally, we compare several variants of our system: purely extractive systems operating over sentences and EDUs respectively, our full system, and ablations removing either the anaphoricity component or parts of the compression module.

In terms of content selection, we see that all of the systems that incorporate end-to-end learning (under ``This work'') substantially outperform our various heuristic baselines. Our full system using the full compression scheme is substantially better on ROUGE than ablations where the syntactic or discourse compressions are removed. These improvements reflect the fact that more compression options give the system more flexibility to include key content words. Removing the anaphora resolution constraints actually causes ROUGE to increase slightly (as a result of granting the model flexibility), but has a negative impact on the linguistic quality metrics.

On our linguistic quality metrics, it is no surprise that the sentence prefix baseline performs the best. Our sentence-extractive system also does well on these metrics. Compared to the EDU-extractive system with no constraints, our constrained compression method improves substantially on both linguistic quality and reduces the number of unclear pronouns, and adding the pronoun anaphora constraints gives further improvement. Our final system is approaches the sentence-extractive baseline, particularly on unclear pronouns, and achieves substantially higher ROUGE score.

\subsection{RST Treebank}
\label{sec:rst}

We also evaluate on the RST Discourse Treebank, of which 30 documents have abstractive summaries. Following \newcite{HiraoEtAl2013}, we use the gold EDU segmentation from the RST corpus but automatic RST trees. We break this into a 10-document development set and a 20-document test set. Table~\ref{table:rst_test} shows the results on the RST corpus.  Our system is roughly comparable to Tree Knapsack here, and we note that none of the differences in the table are statistically significant. We also observed significant variation between multiple runs on this corpus, with scores changing by 1-2 ROUGE points for slightly different system variants.\footnote{The system of \newcite{YoshidaEtAl2014} is unavailable, so we use a reimplementation. Our results differ from theirs due to having slightly different discourse trees, which cause large changes in metrics due to high variance on the test set.}

\begin{table}[t]
\begin{center}
\renewcommand{\tabcolsep}{1.5mm}
\begin{tabular}{|r|rr|} \hline
 & ROUGE-1 & ROUGE-2 \\ \hline
First $k$ words & 23.5 & 8.3 \\ 
Tree Knapsack    & 25.1 & 8.7 \\
Full      & 26.3 & 8.0 \\ \hline
\end{tabular}
\end{center}
\caption{\label{table:rst_test} Results for RST Discourse Treebank \protect\cite{CarlsonEtAl2001}. Differences between our system and the Tree Knapsack system of \protect\newcite{YoshidaEtAl2014} are not statistically significant, reflecting the high variance in this small (20 document) test set.
}
\end{table}

\section{Conclusion}
\label{sec:conclusion}

We presented a single-document summarization system trained end-to-end on a large corpus. We integrate a compression model that enforces grammaticality as well as pronoun anaphoricity constraints that enforce coherence. Our system improves substantially over baseline systems on ROUGE while still maintaining good linguistic quality.

Our system and models are publicly available at \texttt{http://nlp.cs.berkeley.edu}

\section*{Acknowledgments}

This work was partially supported by NSF Grant CNS-1237265 and a Google Faculty Research Award. Thanks to Tsutomu Hirao for providing assistance with our reimplementation of the Tree Knapsack model, and thanks the anonymous reviewers for their helpful comments.

\bibliographystyle{acl2016}
\bibliography{ksumm}

\end{document}